\documentclass[conference]{IEEEtran}
\IEEEoverridecommandlockouts
\usepackage{color, colortbl}
\usepackage{cite}
\usepackage{amsmath,amssymb,amsfonts}
\usepackage{algorithmic}
\usepackage{graphicx}
\usepackage{textcomp}
\usepackage{xcolor}
\usepackage{makecell}
\usepackage{float}
\usepackage{caption}
\usepackage{lipsum}

\makeatletter
\newcommand{\linebreakand}{%
  \end{@IEEEauthorhalign}
  \hfill\mbox{}\par
  \mbox{}\hfill\begin{@IEEEauthorhalign}
}
\makeatother

\def\BibTeX{{\rm B\kern-.05em{\sc i\kern-.025em b}\kern-.08em
    T\kern-.1667em\lower.7ex\hbox{E}\kern-.125emX}}
\begin{document}

\title{Oversampling techniques for predicting COVID-19 patient length of stay\\}

\author{
\IEEEauthorblockN{Zachariah Farahany}
\IEEEauthorblockA{\textit{Department of Computer Science} \\
\textit{Marquette University}\\
zachariah.farahany@marquette.edu}
\and
\IEEEauthorblockN{Jiawei Wu}
\IEEEauthorblockA{\textit{Department of Computer Science} \\
\textit{Marquette University}\\
jiawei.wu@marquette.edu}
\linebreakand
\IEEEauthorblockN{K M Sajjadul Islam}
\IEEEauthorblockA{\textit{Department of Computer Science} \\
\textit{Marquette University}\\
sajjad.islam@marquette.edu}
\and
\IEEEauthorblockN{Praveen Madiraju}
\IEEEauthorblockA{\textit{Department of Computer Science} \\
\textit{Marquette University}\\
praveen.madiraju@marquette.edu}
}

\maketitle

\begin{abstract}
COVID-19 is a respiratory disease that caused a global pandemic in 2019. It is highly infectious and has the following symptoms: fever or chills, cough, shortness of breath, fatigue, muscle or body aches, headache, the new loss of taste or smell, sore throat, congestion or runny nose, nausea or vomiting, and diarrhea. These symptoms vary in severity; some people with many risk factors have been known to have lengthy hospital stays or die from the disease. In this paper, we analyze patients' electronic health records (EHR) to predict the severity of their COVID-19 infection using the length of stay (LOS) as our measurement of severity. This is an imbalanced classification problem, as many people have a shorter LOS rather than a longer one. To combat this problem, we synthetically create alternate oversampled training data sets. Once we have this oversampled data, we run it through an Artificial Neural Network (ANN), which during training has its hyperparameters tuned by using bayesian optimization. We select the model with the best F1 score and then evaluate it and discuss it.
\end{abstract}


%Official IEEE footnote
% \blfootnote{978-1-6654-8045-1/22/\$31.00 ©2022 IEEE}

\section{Introduction}
COVID-19 is defined by the Centers for Disease Control and Prevention (CDC) as "a respiratory disease caused by SARS-CoV-2, a coronavirus discovered in 2019. The virus spreads mainly from person to person through respiratory droplets produced when an infected person coughs, sneezes, or talks"\cite{c1}. Furthermore, they add, "For people who have symptoms, illness can range from mild to severe. Adults 65 years and older and people of any age with underlying medical conditions are at higher risk for severe illness"\cite{c1}.In 2019 this novel coronavirus was first detected. The highly infectious nature of this disease, combined with the respiratory nature of the infection, caused a pandemic. Along with being highly contagious, COVID-19 also has an extensive range of symptoms such as fever or chills, cough, shortness of breath, fatigue, muscle or body aches, headache, the new loss of taste or smell, sore throat, congestion or runny nose, nausea or vomiting, and diarrhea\cite{a3}. Along with a long list of symptoms, COVID-19 has many risk factors, which may increase the severity of the infection. These risk factors include but are not limited to: heart disease, cancer, COPD, obesity, smoking, high blood pressure, and asthma\cite{a2}. With this high infection rate, many symptoms, and risk factors, COVID-19 quickly became a problem for the populations which suffered from it.

\subsection{United States COVID-19 Epidemic}
In the United States, the pandemic began in march of 2020. Since then, the country has mainly recovered, and fewer restrictions have followed. However, there have been over 80 million cases and almost a million deaths in the United States alone\cite{a1}. A significant impact on a developed country is rare in the modern era. Citizens of the United States are particularly at risk because of the high incidence of COVID-19 risk factors such as obesity \cite{a5}. These facts about the American population and COVID-19's infectivity rate caused concern. Experts were concerned that this could cause an upsurge in demand for hospital beds to the point where the healthcare system would get overwhelmed \cite{a7}. In such a case, there would not be enough medical equipment to treat every patient with COVID-19 adequately. This drop in care quality for each COVID-19 patient would then cause an unnecessary increase in the morbidity rate of COVID-19. This is avoidable given the proper system, which optimizes patient flow, a statistic which indicates proper hospital management.

These drastic health concerns caused concern amongst US legislators, and they quickly implemented stay-at-home orders. Many theorized that these orders lessened the pandemic's severity by keeping large groups of people from gathering. The cost of easing the pandemic was a loss in economic output because workers were obligated to stay home away from their place of work\cite{a4}. This situation caused economic output to drop since remote employment was only beginning as a viable option for employers, and some jobs could not be performed remotely. Nevertheless, the United States and other countries worldwide suffered through this period of economic hardship and disease. The advent of vaccines for COVID-19 caused a return to normalcy and a subsequent reopening of the economy. Nevertheless, it remains crucial to detect people at risk for severe COVID-19 infections early for the sake of the healthcare system and the populace.

Due to the healthcare and economic significance of appropriately managing COVID-19 patients, it is critical to have a good prediction of their infection's severity beforehand. Such a method would allow healthcare workers to better address individual patients who might require more attention. The entire healthcare system would also be more efficient because it would be evident which cases are high-severity. The hospital could then allocate more resources to these high-severity cases. Without such a system, it would be harder to know where to assign healthcare resources, and hospital efficiency would drop. If this were to happen, overwhelming the healthcare system would be more likely causing unnecessary deaths. We propose using deep learning to predict patient severity through the length of stay (LOS) metric. LOS is a common and important metric. It indicates the severity of a patient's infection, overall patient health, and hospital efficiency. This project aims to predict a COVID-19 patients LOS using information available through hospital databases. A patient's Electronic Health Records (EHR) provide the data necessary to make such a prediction accurately.

\subsection{Wisconsin and COVID-19}
\par
Upon investigation, Wisconsin has had its own unique economic, social, and medical problems because of COVID-19. Within a study by Malecki et al., 42\% of the respondents indicated a change in their work due to COVID-19. Also 32\% of female and 27\% of male respondents reported a decrease in wages during the onset of the pandemic \cite{b5}. These statistics show the overall lack of economic stability in Wisconsin because of the outbreak of COVID-19. They also provide a motivator to attempt to lessen the severity of COVID-19 and future pandemics.
\par
From this study, it also appears that the social and psychological aspects of people's lives are being harmed as well. $>$80\% of participants reported canceling social gatherings due to COVID-19\cite{b5}. Also, it seems that COVID-19 has caused participants to report increased stress especially if the participants have children which are younger than 18\cite{b5}. This shows that COVID-19 not only directly affects peoples physical health, but it has affected the general levels of stress across families in Wisconsin.
\par
Lastly, it appears that people's experiences with healthcare have changed due to the pandemic. Many experienced a lack of access and delays in healthcare during the pandemic\cite{b5}. The most common reason for this was reported to be postponing or cancellation due to COVID-19. 44\% reported voluntarily delaying their medical care due to fears of contracting COVID-19. From the data and discussion provided by Malecki et al., it appears that the healthcare quality that the average citizen receives in Wisconsin has gone down. As previously stated, this could be due to cancellations, delays, or even voluntarily. However, this will affect the Wisconsin population. It also provides an increased motivation to mitigate the further spread of COVID-19 and optimize hospital systems.
\par
The rest of this paper is organized as follows. In section II, we discuss the related work which informed our research methods. Section III covers the methods for obtaining, processing, and modeling the data. In section IV, we share our results and provide a discussion of them. Section V is our conclusion, where we discuss the significance of our findings, potential applications, future work, and limitations. All of the work in this project is subject to an IRB exemption. This is because the project uses de-identified patient data with all dates subject to random offsets.
\section{Related work}
\par
The prediction of LOS is not a new task; many have attempted it in various ways. Daghistani et al. used a data pipeline for cardiac patients where they selected features based on the information gain statistic\cite{b1}. After, they use SMOT to balance the classes of the training set. Lastly, they compiled various models, evaluated them, and selected the best one for the task. They created a Random Forest (RF) model, which achieved 80\% accuracy and an 80\% F1 Score. They also created an ANN which achieved 50\% accuracy and 49\% F1 score. These statistics were all created through 10-fold cross-validation. The advantage to their method\cite{b1} is that information gain statistic was used to have their models parameterized by only the top 20 most important features. This makes it easy for them to input patient information and receive an estimate of the patient's LOS as a discrete variable.
\par
Another study instead treated the problem as a regression task in which their predicted variable is continuous. Mekhaldi et al. used an open-source Microsoft dataset to try to predict LOS. Their methods are unique because they used SMOTER, which reduces the imbalance of a continuous variable. They argue that the prediction of a continuous variable is more valuable to the hospital system \cite{b2}. This is because a continuous variable can have extreme values, which the model must learn. In the discrete case, the model only learns the distinction between categories, and the high severity category accounts for all extremely high values. This is not the case in the continuous case where the model will be penalized for not learning when to predict a case to be very severe. This is another reason the authors decided to use SMOTER. SMOTER, in this case, was used to create more artificial samples at the high end of the LOS distribution. This allows their model to better learn how to predict severe cases.
\par
The effectiveness of machine learning was demonstrated in a study on sepsis\cite{b3}. Shimabukuro et al. exhibited the effectiveness of an experimental machine learning algorithm relying on common EHR data to provide early predictions of sepsis. This model was being compared to a traditional rule-based sepsis prediction. The traditional approach was used on the control group (n = 75) and the machine learning model was used on an experimental group (n = 67). When either method decided a patient had severe sepsis, the patient began treatment accordingly. Their method beat the traditional sepsis detection approach and significantly improved patient outcomes. The control group had a mortality rate of  21.3\% and the experimental group had a mortality rate of  8.96\% which is statistically significant (p$<$.05). The ICU LOS and hospital LOS also decreased in a statistically significant manner. The effectiveness of the machine learning model within this study shows how early detection of diseases and severity is imperative in a hospital setting. Shimabukuro et al. proved that such models could significantly impact patient outcomes and that machine learning models can beat traditional methods.
\par
Real-time models which try to predict patient discharge at 2 pm or midnight instead of LOS are helpful in a clinical context\cite{b4}. Essentially these models make a prediction for each patient each day. Barnes et al. showed that by using a Random Forest or Logistic regression, a hospital can beat the sensitivity of clinicians predictions. These methods did, however, not outperform clinicians in terms of specificity. This indicates that the machine learning methods were better at identifying cases where the patients were likely to be discharged. However the clinicians were better at predicting patients who would not be discharged. The author also discussed why proper hospital resource allocation is crucial and relies on solid predictions. Another way to conceptualize optimal use of hospital resources is through patient flow which is an essential indicator of "patient safety, satisfaction, and access."\cite{b4} The results of this paper are promising because they demonstrate the possibility of machine learning models working in conjunction with doctors to create real-time results.
\par
Similar studies of predicting LOS using machine learning approaches have also been implemented on COVID-19 data. Dina et al.~\cite{alabbad_machine_2022} discretized data into nine cohorts based the LOS and the data set were imbalanced. They used SMOTE to remedy this. After which, they employed a plethora of machine learning algorithms, RF  yielded the best result with 94.16\% accuracy. Moreover, the authors used this model to get the essential features that impacted LOS, such as age, CRP, and nasal oxygen support days. In another study, the authors just discretized the data of COVID-19 patients in the emergency department (ED) into two groups according to their stay hours (i.e., LOS $<$4 hours or $\ge$ 4 hours) ~\cite{etu_prediction_2022}. There were 127 clinical and operational variables in the original dataset. They used the Kolmogorov–Smirnov (K-S) test to identify the dataset's correlation and eliminate less significant features. Only 60 features were selected. Similarly, they used SMOTE to fix the imbalance within the data before creating machine learning models. Studies of purposes, the models and algorithms, data source, input features and results are summarized in Table~\ref{table:RelatedWorks}.
\par
Even though using machine learning in a hospital context to predict LOS or discharge dates is common, our approach goes in a new direction. Our extensive oversampling and hyperparameter tuning provide a unique way to create the best performing models for predicting patient LOS. We hope our work will push further research into this area, as many imbalanced classification problems exist in the medical world, like predicting LOS. Such work could include more advanced oversampling techniques or a better technique for gathering the feature importances from a neural network.

\begin{table*}[!htbp]
\caption{Related papers}
\begin{center}
\begin{tabular}{c c c c c c c}
\hline
\textbf{Ref} & \textbf{Aim} & \textbf{Model} & \textbf{Methods } & \textbf{Data-set} & \textbf{Model inputs} & \textbf{Results} \\
\hline
\makecell{[9]} & \makecell{Predict in-hospital \\ LOS for cardiac \\ patients by ML \\ based model.} & \makecell{ML models} & \makecell{RF, ANN,\\ SVM, BN} & \makecell{12,769 unique \\ patients’16,414 \\visits  between \\ 08 to 16 at \\ KACC, Riyadh. }  & \makecell{Demographic, cardiovascular \\risk factors, admission\\ and discharge diagnosis, \\vital signs, and laboratory \\ tests on admission info \\ extracted from EMR.} & \makecell{F Score: \\ RF 80\% \\BN 50\% \\SVM 67\% \\ANN 49\%
}\\

\makecell{[10]} & \makecell{Implement ML \\ process for LOS \\ prediction.} & \makecell{ML models} & \makecell{RF, GBM} & \makecell{Microsoft data set\\for LOS prediction}  & \makecell{one-hot-encoding to \\represent categorical data} & \makecell{MAE:\\RF 0.44\\GBM 0.55\\$R^2$ / Adjusted $R^2$:\\RF 0.92\\GBM 0.91}\\

\makecell{[11]} & \makecell{Tested ML based \\severe sepsis \\ prediction system \\for reductions in\\ average LOS \\and in-hospital \\mortality rate. } & \makecell{ML models, \\ Disease severity \\ scoring system} & \makecell{Machine learning \\ algorithm (MLA) \\ SOFA, qSOFA,\\SIRS, MEWS} & \makecell{Randomised clinical \\ trial in two mixed \\ICUs at the \\ UCSF Medical Center \\at Parnassus Heights \\ from Dec ‘16 \\ to Feb ‘17.}  & \makecell{-} & \makecell{\\AUROC:\\MLA 0.952\\SIRS 0.681 \\ MEWS 0.524\\ SOFA 0.756\\ qSOFA 0.518\\Sensitivity:\\MLA 0.900\\SIRS 0.590\\MEWS 0.365\\SOFA 0.910\\qSOFA 0.288\\
}\\

\makecell{[12]} & \makecell{Automate and \\improve patient\\ discharge \\predictions by \\supervised ML \\methods.} & \makecell{ML models} & \makecell{RFF, LR} & \makecell{Patient flow data (i.e.,\\ admission and \\discharge times), \\demographics, and \\basic admission \\diagnoses data  \\from January 1,\\ 2011 to November \\1, 2013 of \\9636 patient visits.}  & \makecell{Standardized data\\ (Except numerical \\variables all \\remaining predicators \\are modeled \\using binary)} & \makecell{Sensitivity (\%)\\2 p.m\\ LR 65.9 \\ RRF 60.0\\End of Day\\LR 71.5\\RRF 66.1
}\\

\makecell{\cite{alabbad_machine_2022}} & \makecell{Implement ML \\process for \\LOS prediction \\of COVID-19 \\patients and \\identify the\\ most relevant \\features affecting\\ LOS.} & \makecell{ML models} & \makecell{RF, GB, \\XGBoost, \\Ensemble Models} & \makecell{895 COVID-19 \\patients admitted\\ to King Fahad \\University hospital \\in the eastern \\province of\\ Saudi Arabia.}  & \makecell{Clinical and\\ demographics data\\ contain comorbidities,\\ laboratory results\\ symptoms, and \\vital signs,\\ 47 features\\ in total.} & \makecell{Accuracy (\%) \\RF: 94.16\%
}\\

\makecell{\cite{etu_prediction_2022}} & \makecell{Predict LOS \\of COVID-19 \\patients in the\\ emergency \\ using ML models.} & \makecell{ML models} & \makecell{LR, GB,\\ DT, RF} & \makecell{3,301 COVID-19\\ patients in \\Detroit from \\March 16 to \\December 29, '20.}  & \makecell{Normalized and \\selected features\\ by statistical \\approaches such\\ as K-S test, \\a one-way analysis\\ of variance \\(ANOVA) and \\Spearman correlation.} & \makecell{Accuracy (\%)\\GB: 85\\DT: 82\\RF: 82\\F1:\\GB: 88\\DT: 86}\\
\hline
\multicolumn{7}{l}{ \makecell{ Random Forest (RF), Artificial Neural Network (ANN), Support Vector Machine (SVM), Bayesian Network (BN), Gradient Boosting \\Model (GBM), Sequential Organ Failure Assessment (SOFA), quick SOFA (qSOFA), Systemic Inflammatory Response Syndrome (SIRS),\\ Modified Early Warning Score (MEWS), Regression Random Forest (RRF), Logistic Regression (LR), Extreme Gradient Boosting (XGBoost), \\Decision Tree (DT), Kolmogorov–Smirnov test (K-S test), King Abdulaziz Cardiac Center (KACC)} }

\end{tabular}
\label{table:RelatedWorks}
\end{center}
\vspace{-4mm}
\end{table*}

\section{Methods}
\subsection{Data Source}This data was granted by the Froedert Hospital in Wauwatosa Wisconsin. The files granted were diagnosis, diagnostic results, encounters, medication orders, medications administered, naaccr, patient demographics, problem list, procedures, social history lifestyle, and vitals. All information in these tables was subject to random time offsetting so that the patients would remain anonymous. Patient names and identifiers were also removed. The de-identification process is standard for use in medical data and is necessary to ensure the patient's privacy. It is also necessary to ensure consensual data usage. The data was compiled on June 2nd, 2021, and the data goes back to June 4th, 2016. This data was then combined with additional data containing past patient substance use. This second data set was only used to supplement the first in hopes that they may share information on the same set of patients.

\begin{figure}
    \centering
    \includegraphics[width = 8cm]{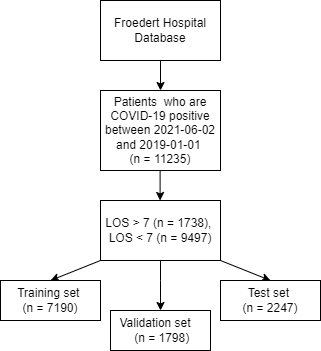}
    \caption{Data pipeline}
    \label{fig:datapipeline}
\end{figure}

\definecolor{Gray}{gray}{0.9}
\begin{table*}[!htbp]
    \centering
    \caption{Descriptive statistics of the training set}
    \begin{tabular}{c c c c c c}
        \rowcolor{Gray}
        \textbf{Feature}& \textbf{Total population} & \textbf{$<$30 day LOS} & \textbf{30 day $<$ LOS $<$ 60 day} & \textbf{60 day $<$ LOS $<$ 90 day} & \textbf{$>$90 day LOS}\\
        n & 7190 & 7061 & 94 & 17 & 10 \\
        Median age at visit(years) &54.11&53.91&60.83&56.26&65.97 \\
        Median weight at visit(lbs) &185&185&190.18&185&165.12 \\
        Median height at visit(inches) &66&66&65.99&67.99&65.24 \\
        Median BMI at visit(kg$/$m$^2$) &29.65&29.65&29.47&29.22&26.76 \\
        \rowcolor{Gray}
        Sex &&&&&\\
        Men & 2422 & 2347 & 55 & 11 & 6\\
        Women & 4768 & 4714 & 39 & 6 & 4\\
        \rowcolor{Gray}
        Race &&&&&\\
        
        White& 5036 & 4961 & 54 & 12 & 5\\
        
        Black& 1664 & 1623 & 31 & 5 & 3\\
        
        Other/not specified& 490 & 477 & 9 & 0 & 2\\
        \rowcolor{Gray}
        Ethnicity &&&&&\\
        
        Non-Hispanic&6840&6725&83&17& 9\\
        
        Hispanic&342&328&11&0& 1\\
        
        Other/not specified&8&8&0&0&0 \\
        \rowcolor{Gray}
        Language &&&&&\\
        
        English&7028&6906&88&17&9 \\
        
        Spanish&87&82&5&0&0 \\
        
        Other/not specified &75&73&1&0&1 \\
        \rowcolor{Gray}
        Marital Status &&&&&\\
        Married &3830&3774&42&11&2\\
        Single &2030&1989&31&4&3\\
        Divorced &573&553&14&1&4\\
        Other/not specified &757&745&7&1&1\\
        \rowcolor{Gray}
        Employment &&&&&\\
        Full time &2857&2815&31&11&0\\
        Not Employed &901&887&14&0&0\\
        Retired &2060&2016&30&3&5\\
        Part time &535&533&1&0&1\\
        Other/Not specified &837&810&18&3&4\\
    \end{tabular}
    \rule{0pt}{1ex}
    \label{table:stats}
\end{table*}

\subsection{Data Processing}
Once the data was collected, it was essentially a collection of hospital entries about hospital-to-patient interactions. These were separated under what type of interaction took place. The same applies to all the files analyzed in this data set. Taking the diagnosis file as an example, this file can contain multiple diagnoses and thus multiple rows which apply to the same patient. Since we are trying to predict the LOS for each patient, this format will not suffice. We want to have patients and all the data which applies to them in one row. To accomplish this, a pivot table was created to hold all patient IDs. This pivot table was then populated with all of the information from the diagnosis file. This same process was conducted on all files within the data set. This same process was conducted on all of the files within the data set. Once we had a pivot table for each file, we conducted an inner merge of all of the separate pivot tables.
\par
The diagnosis file contained each patient's number and an ICD-11 code which pertains to an individual diagnosis. This was useful for ascertaining any comorbidities the patients might have. Medications ordered and medications administered contained any information available about what medications each patient was using. Patient demographics contain information on the marital status, race, ethnicity, language, sex, and age of each patient. The problem list file contains further diagnoses. The procedures table contains any procedure like a surgery or an x-ray the patient received during their stay. Social history lifestyle contained information on the patient's drug and alcohol use. Vitals had information on the patients height, weight, and BMI. Lastly the encounters file contains the department which the patient interacted with, the admission date, and the discharge date.
\par
The encounters file received extra processing steps. We used the admission date to calculate the day of the year from 1-365. This field was meant to capture any seasonality in the data set. We also used the admission date to calculate the ordinal time since 1970. This column was meant to represent any ordinal relationship in the data set. The admission and discharge times were also used to yield the LOS as a continuous variable. We chose to discretize this variable into two classes. Cases that are not severe $<$7 days, and cases that are $>$7 days.
\par
As our data set was very large, sparse, and contained many missing values, we removed any column which has more than 50\%. missing values. We also removed any column that had zero variance. This would indicate that a column has the same values and therefore provides no information. As there were not a lot of missing values remaining, they were imputed using K nearest neighbors imputation. We added indicators to the imputed data to show which cells contained imputed data. These indicators were left as columns in the training set so the network could learn which data was imputed if necessary.

\subsection{Final Dataset}
At the end of our data processing, we will conduct a 80-20 stratified train test split. This means that we will split our data set into train and test sets, ensuring that they have the same amount for each class. We then conduct another 80-20 stratified split to get our validation data. Now we have three separate data sets. This process is visualized in Fig.~\ref{fig:datapipeline}.
\par
Our training data consists of 7190 rows and 25008 columns. This resulting data is extremely sparse as most of the entries are zero or one. We show the descriptive statistics of some of the variables in our data set in Table~\ref{table:stats}.
\par
\begin{figure}[H]
    \centering
    \includegraphics[width=10cm] {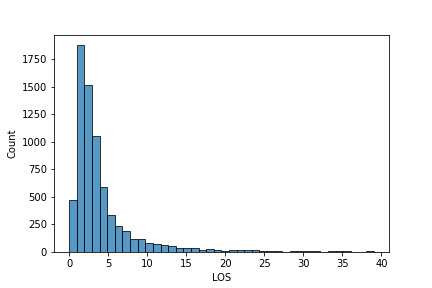}
    \caption{Distribution of the LOS}
    \label{fig:loshist}
\end{figure}

Most of the patients in the data set had a LOS which was $<$30 days. This is the source of the the imbalance in our classification problem. This imbalance is visualized in Fig. \ref{fig:loshist}. Another problem with this data set is the bias that is present within it. This data includes patients that are majority older, white, English speaking, and non-Hispanic. This means that our models will see more samples that fit this description. Since this description is prevalent in the data, our model can perform better predictions for this type of person. Conversely, the model will not be as good at making predictions for people who do not fit this description. However, such a problem is very likely when gathering data from a hospital in Wisconsin because of the state's demographics.

\begin{figure*}
    \centering
    \includegraphics[width=18cm] {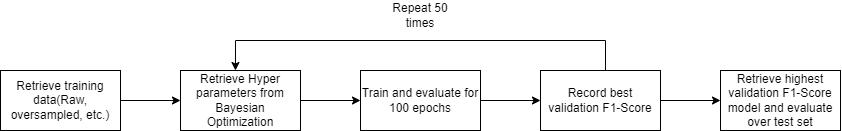}
    \caption{Training pipeline}
    \label{fig:trainingloop}
\end{figure*}

\begin{figure}[H]
    \centering
    \includegraphics[width=10cm] {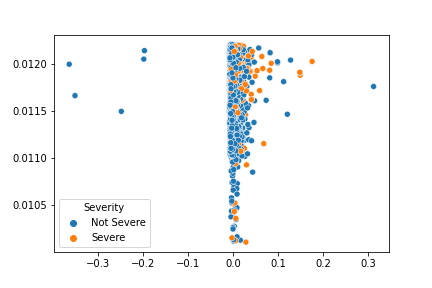}
    \caption{Two component PCA with classes included}
    \label{fig:pca}
\end{figure}

Due to the high dimensionality of our data, it can be difficult to visualize. To alleviate this, we conducted Principal Component Analysis (PCA) on the continuous data points in order to project it onto two dimensions where it could be more easily visualized. This visualization is shown in Fig.~\ref{fig:pca}. We can see that it appears very hard to visually separate the data points to classify them.

\subsection{Modeling pipeline}
Given such a large body of data and frequent use in previous work, we use ANNs to predict the LOS as a binary categorical variable. The primary metric which we will be using is the F1 score and this is because we are dealing with an imbalanced dataset. The number of low LOS patients to high is about 5-1 or an 80/20 split, making achieving a high accuracy trivial. A model could predict only low LOS and would reach 80\% accuracy. F1 Score is a better metric for an imbalanced problem because it is the harmonic mean of precision and recall. Also, achieving a high F1 score in an imbalanced class problem is less trivial than achieving a high accuracy making it more useful for evaluation. If we can achieve a high F1 score, our model has learned the underlying relationships in the data.
\par
In addition, to maximize our effectiveness in combating this imbalanced class problem, we will create five separate training data sets over which we will evaluate training. These data sets will reflect various oversampling techniques that have been shown to assuage imbalanced class problems and could increase the F1 Score. The first data set will be the raw training data. The second will be the raw data set in conjunction with attaching a weight to high LOS instances in our loss function. This should make the loss associated with positive vs. negative cases equal. The third training data set was constructed by using random oversampling on the original training data set. Similarly, the fourth data set was created using random undersampling until the training classes were even. Lastly, the final training data set was obtained using SMOTE-NC, which makes new instances of the minority class by getting random samples of the minority class and finding an arbitrary point between the samples that would constitute a synthetic minority example.
\par
ANNs can vary in many hyperparametric ways. For instance, the number of layers in a network or the number of neurons in each layer. In order to find the optimal hyperparameters, we define a sweep using the hyperparameters listed. Finding optimal hyperparameters is crucial to the performance of our model and how well it generalizes.

\begin{table}[ht]
    \centering
    \caption{Hyperparameter choices}
    \begin{tabular}{|c|c|}
        \hline
        \textbf{Hyperparameter} & \textbf{Values} \\
        \hline
        Optimizer & Adam, Nadam, Adamax,\\
         &Rmsprop(momentum = .9), \\
         &Adagrad, Adadelta\\
        \hline
        Number of layers & 1-5, step size = 1 \\
        \hline
        Number of neurons per layer & 1000-5000, step size = 500 \\
        \hline
        Dropout rate & 0-.5, step size = .1\\
        \hline
        Learning rate & 1e-8 – 1e-2, logarithmic step sizes\\
        \hline
        Regularization type & L1, L2, L1 and L2\\
        \hline
        Regularization factor & 1e-8 – 1e-2, logarithmic step sizes\\
        \hline 
        Activation function & ReLU, LeakyReLU, eLU, SeLU,\\
         & PReLU, GeLU, Swish\\
        \hline
        Alpha value (For LeakyRelU) & 0-.5, continuous \\
        \hline
    \end{tabular}
    \hspace{1cm}
    \label{tab:hyperparam_choices}
\end{table}

The sweeping method which was used is Bayesian optimization\cite{BayesOpt}. Bayesian optimization uses Bayes' theorem to, in essence, “zoom in” on areas of the hyperparameter space where better F1 Scores are more likely. For Bayesian optimization to work, the user must specify an objective. In our case, this is the validation F1 score. Bayesian optimization works by creating a surrogate function that is meant to model the relationship between the hyperparameters and the objective. The inputs of the surrogate function are hyperparameters, and the output is the expected performance. With a sufficiently accurate surrogate function, one can inexpensively find the global maximum of the surrogate function. This maximum is the next set of hyperparameters to train on and evaluate. For our search, we use the validation set to calculate the validation F1 Score. We iterate this searching method and search 50 times to locate the hyperparameters which produce the best validation F1 Score. The choices for these hyperparameters are listed in Table~\ref{tab:hyperparam_choices}. After this we use the best-performing model to calculate the test F1 Score. We also include other metrics which were gathered for completeness and a more thorough analysis of their performance.

\section{Results}

\begin{table*}[ht]
    \centering
    \caption{Hyperparameters selected through Bayesian optimization over 50 runs}
    \begin{tabular}{|c|c|c|c|c|c|}
        \hline
        & Model \#1 & Model \#2 & Model \#3 & Model \#4 & Model \#5\\
        \hline
        Hyperparameters &  Raw training & Raw training & Random oversampled & Random undersampled & SMOTE-NC\\
         & data& data with weights& training data& training data& training data\\
        \hline
        Optimizer & Adamax & Adam & Rmsprop & Adagrad & Adamax\\
        \hline
        Number of layers & 3 layer & 1 layer& 1 layer& 1 layer& 3 layer\\
        \hline
        Number of neurons & 4000 neurons & 2000 neurons& 2000 neurons& 2500 neurons& 4000 neurons\\
        per layer &&&&&\\
        \hline
        Dropout rate & 0.3 & 0.5 & 0.4 & 0.1 & 0.3\\
        \hline
        Learning rate & 1e-5 & 5e-6 & 1e-7 & 1e-3 & 1e-5\\
        \hline
        Regularization type & L1 and L2 & L2 & L1 and L2 & L2 & L1 and L2\\
        
        \hline
        Regularization factor & 5e-5 & 1e-3 & 1e-7 & 5e-8 & 5e-5\\ 
        \hline
        Activation function & Leaky ReLU & eLU & Swish & eLU & ReLU\\
        \hline
        Alpha value & .04 & NA & NA & NA & NA\\
        \hline
    \end{tabular}
    \hspace{4cm}
    \label{tab:hyperparam_results}
\end{table*}

As visualized in Table~\ref{tab:hyperparam_results}, most selected models tend to be shallow and wide rather than deep. This could be because deeper models are much more complex and, therefore, more prone to overfitting. Since we used the validation set to select the best-performing model, overfitting would be problematic because it would lead to a model having low validation scores. Shallow models appear to be optimal in order to generalize from the training set to the test set.

\begin{table*}[ht]
    \centering
    \begin{tabular}{|c|c|c|c|c|c|c|}
        \hline
        \textbf{Model} & \textbf{Data trained on} & \textbf{F1} & \textbf{Accuracy} & \textbf{Precision} & \textbf{Recall} & \textbf{AUC} \\
        \hline
        Model \#1 & Raw data & \textbf{85.79}\% & \textbf{95.50}\% & 86.64\% & \textbf{84.95}\% & \textbf{91.23}\%\\
        \hline
        Model \#2 & Raw data + weights & 79.11\% & 92.47\% & 90.90\% & 70.02\% & 84.11\%\\
        \hline
        Model \#3 & Random Oversampled & 27.26\% & 50.37\% & 59.37\% & 17.69\% & 52.14\%\\
        \hline
        Model \#4 & Random Undersampled & 69.62\% & 87.18\% & \textbf{93.75}\% & 55.36\% & 77.02\%\\
        \hline
        Model \#5 & SMOTE-NC & 79.15\% & 92.57\% & 90.05\% & 70.60\% & 84.33\%\\
        \hline
    \end{tabular}
    \caption{Performance of each model}
    \hspace{10cm}
    \label{tab:model_performance}
\end{table*}

\begin{figure}
    \centering
    \includegraphics[width = 10cm]{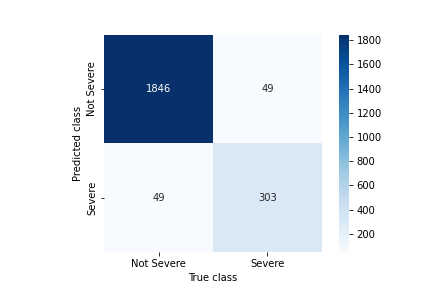}
    \caption{Confusion Matrix Model \#1}
    \label{fig:confmat}
\end{figure}

From our results in Table~\ref{tab:model_performance}, we can see that the pipeline which used the raw training data outperformed the competition in almost every category. The only category which model \#1 did not perform the best is precision. In this category, it was beaten by three of the other four models. Furthermore, since model \#1 performed the best on almost all of the metrics is not the most surprising because it was trained on data that is most similar to the testing data. It is unclear whether precision would be essential in a hospital setting. If one employs a low precision model, many false positives could impede optimal patient flow. However, due to this model's overall performance, we think it is the best choice to implement within a hospital. It surpasses the other models in all aspects besides precision and it still retains a very high precision score. We made this choice because it is simply impossible to recommend switching to any of the other models because they all have serious flaws. Model \#1 has the flaw of a lower precision score, but it still has the best-case precision-recall trade-off. The results of model \#1 are shown in Fig.~\ref{fig:confmat}.

\par
Model \#4 which uses random under sampling performed the best in precision. This means that model \#4 predicted an instance as severe had a high chance of being correct. In our use case, this could be considered useful if one wanted the model to be more conservative. Such a model would not find all severe cases but if it predicted a patient as severe, it is likely to be correct. In a hospital setting, if such a model were deployed, it would be more proficient in avoiding false positives. Such a situation where a hospital would want a high precision is conceivable, but it is more likely that they would want a higher recall. The recall in this model was severely underperforming the rest. Such a low recall indicates that the model did not classify many severe cases as severe. In the context of a hospital, the model missed many severe patients. Under certain circumstances, this could cause resource mismanagement and sub-optimal patient flow. This is because it might cause the medical staff to pay less attention to people who were wrongly classified as not severe. 
\par
If we were to recommend a model as an alternative to model \#1 to have higher precision, it would be model \#5. This is because model \#5 outperforms model \#1 in precision. Moreover, it retains a competitive performance in all other aspects. It suffers a little bit from low recall, which means it will miss some severe cases, but that is unavoidable if we wish to increase precision. Unfortunately, precision and recall have an inverse relationship, so it is difficult to maintain one while increasing the other.
\par

\begin{figure}
    \centering
    \includegraphics[width = 9cm]{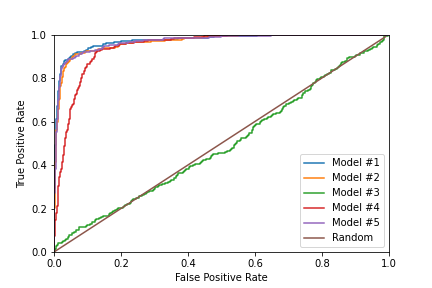}
    \caption{Receiver operator curve(ROC) for all models}
    \label{fig:roc}
\end{figure}

\vspace{2cm}

Also, it appears that model \#3 is the lowest-performing model and is essentially useless. This is because it used undersampling meaning, meaning the model has less data to use in training. Since it had less data, it could not learn the underlying relationship that all of the other models were able to learn. This conclusion is further supported by Fig.~\ref{fig:roc}. This is one of the main drawbacks of random undersampling. If there is a significant difference between the number of majority class instances and minority class instances, then a model could not learn the relationship at all.

\par
\begin{figure}[H]
    \centering
    \includegraphics[width = 9cm]{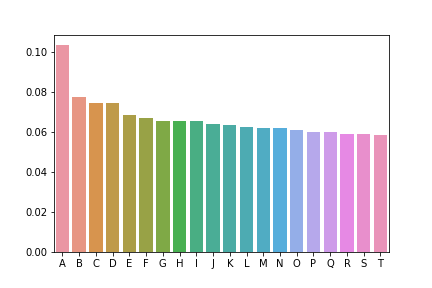}
    \caption{Mutual information scores of the top 20 features}
    \label{fig:featurescores}
\end{figure}

Finally, let us estimate how important each feature was for predicting the LOS. To estimate the importance of each feature, we use the mutual information statistic for classification. We chose to visualize the top 20 most important features in Fig.~\ref{fig:featurescores}. Many of the features displayed are lengthy notes on procedures conducted on the patients, so we included an external reference table in Table~\ref{tab:featurereferencetable} to see the features.

\begin{table}
    \centering
    \begin{tabular}{|l l|}
        \hline
        \multicolumn{2}{|l|}{\textbf{Feature label Full feature description}} \\
        A- &  Subsequent hospital care per day for the evaluation and \\&
        management of a patient which requires at least 2 of these 3 \\&key components. A detailed interval history. A detailed examination 
         \\&medical decision making of high complexity Counseling and \\&or coor.\\
        B- & Critical care evaluation and management of the critically \\ &ill or  critically injured patient first 30-74 minutes\\
        C- & Intravenous infusion for therapy prophylaxis or diagnosis\\ & specify substance or drug additional sequential infusion \\ & of a new drug substance up to 1 hour. List separately in addition\\ & to code for primary procedure\\
        D- & Intravenous infusion for therapy prophylaxis or diagnosis\\ & specify substance or drug each additional hour. List \\ & separately in addition to code for primary procedure.\\
        E- & CULTBLOOD\\
        F- & Gases blood any combination of pHp CO2 pO2 CO2 HCO3\\ & including calculated O2 saturation with O2 saturation by\\ & direct measurement except pulse oximetry\\
        G- & Lactatelacticacid\\
        H- & POCHESTPAORAP\\
        I- & Anticoagulants\\
        J- & Magnesium\\
        K- & Radiologic examination chestsingle view\\
        L- & MAGNESIUMQuantitative\_na\\
        M- & Subsequent hospital care per day for the evaluation and\\ & management of a patient which requires at least 2\\ & of these 3 key components. An expanded problem focused\\ & interval history. An expanded problem focused examination.\\ & Medical decision making of moder.\\
        N- & IP CONSULT FOR VASCULAR ACCESS TEAM\\
        O- & IP CONSULT TO NUTRITIONAL SERVICES\\
        P- & Culture bacterial blood aerobic with isolation and presumptive\\ & identification of isolates includes anaerobic culture if appropriate\\
        Q- & Therapeutic prophylactic or diagnostic injection specify substance\\ & or drug each additional sequential intravenous push of the same\\ & substance drug provided in a facility. List separately in addition\\ & to code for primary procedure.\\
        R- & Therapeutic activities direct one on one patient contact use of\\ & dynamic activities to improve functional performance each \\&
        15 minutes.\\
        S- & MAGNESIUMBLOOD\\
        T- & CULTUREBLOOD\\
        \hline
    \end{tabular}
    \caption{Feature reference table}
    \label{tab:featurereferencetable}
\end{table}

\par
It seems that magnesium levels of the patients were critical in determining the severity of their illness. We draw this conclusion because 3 features in the top 20 pertain to magnesium levels in the blood. Also important seems to be the drawing of a blood culture, which again indicates that blood content may be an important factor in predicting LOS.

\par
Another common trend within the important features seems to be the presence of intravenous infusions for prophylactic reasons. This indicates that the medical staff could tell that a patient would require fluids or medication through an IV. They then administered medication through an IV in anticipation of an illness. Intuitively the use of an IV would make sense to predict longer LOS because if one has an IV in the condition is likely more serious and long-term.

\par

\vspace{4cm}

Lastly, the top two most important features seem to indicate very severe problems with the patient's health. In feature A, they mention "high complexity counseling." In feature B the patient is in critical care within the first 2 hours of care. These two features have a solid relationship to the severity of the COVID-19 infection.

\section{Conclusion}
In short, we have demonstrated the effectiveness of large ANNs in predicting LOS. They can achieve markedly larger accuracy and F1 scores. As we discussed, the proper treatment of COVID-19 patients is especially important since large-scale COVID-19 outbreaks risk overwhelming hospitals. We managed to provide multiple methods for predicting a COVID-19 patient's LOS. The models which we presented had varied effectiveness. Some achieved very high accuracy, favoring the majority class like model \#1, and some had more balanced predictions like model \#2 or model \#5. Any of these options would provide great utility in a hospital setting. Models \#2 and \#5 showed how oversampling could be used to gain more balanced results when dealing with an imbalanced classification problem. Also, through our estimation of feature importances, we provided information on which features help predict the severity of a COVID-19 infection. Many of the discoveries we made in the feature space were novel. Seemingly innocuous features like magnesium levels were very important in predicting severity.
\subsection{Significance}
This study suggests that hyperparameter tuning with deep learning can be used to great effect in a hospital setting. They can identify patterns in patient data that humans cannot and, therefore, can be used to improve hospital management and patient flow. The proper prediction of severity for COVID-19 mitigates the chance of hospitals getting overwhelmed. 
\subsection{Limitations}
\par
One of the major limitations of our study was the homogeneity of the data we used. As previously mentioned, the patients studied were primarily older, white, English-speaking, and non-Hispanic. Our model does not have any information about how to model the LOS of non-white non-English speakers, and it could underperform for people in this group. This, unfortunately, is the reality of using data from one source and is very hard to remedy without collecting more data.
\par
Another limitation was the high dimensional of our data. As previously mentioned, our final training set had 7190 patients and 25008 columns. In this study, we wanted to use the maximal amount of information to feed into our training pipeline. We did this because we are using ANNs, which are known to be data hungry. However, in some respects, this served to be a limitation. This large amount of columns made it computationally impossible to calculate the perturbation feature importance of each model. This is why we had to estimate the feature importance using the mutual information statistic simply. 
\par
Lastly, the primary limitation of our dataset is that it describes patient-to-hospital interactions only within Froedert Medical center. This poses a limitation in two ways: Firstly, our models would not necessarily generalize to other regions outside of Wisconsin. Secondly, the data does not capture medical interactions in other hospitals. For instance, if a patient received a diagnosis in another hospital and did not report that diagnosis to Froedert, we would have no information on it.
\subsection{Future work}
To remedy the problem of homogenous data, we could obtain more comprehensive data from across the United States hospital system. The same methods we used here would work and then be able to accurately predict LOS for a broader range of people from all over the United States.
\par
Another possibility for future work would be more advanced oversampling techniques. A GAN\cite{z1} could be used to generate similar data for our minority class if granted access to the proper computational power. This oversampling method could prove to be vastly superior to the one we demonstrated in this paper. We were unable to implement this idea due to the high computation complexity of GANs.
\par
A second possibility would be to gather the feature perturbation importance from the models. This could be through either reducing the dimensionality of the data or using more powerful hardware. This would allow us to see which features the neural network considered important to predict the LOS. Having feature importances from the model would make it easier to interpret.
The analysis in the project was conducted using Python, Sklearn, Keras, Tensorflow\cite{t1,t2,t3}. Lastly, Weights and Biases was used to conduct hyperparameter tuning and hold validation model results\cite{t4}.

\section{Acknowledgement}
The first author was supported by the National Science Foundation grant \#1950826.

\bibliographystyle{IEEEtran}
\bibliography{bibfile}
\end{document}